\documentclass[lettersize,journal]{IEEEtran}
\usepackage{amsmath,amsfonts}
\usepackage{algorithmic}
\usepackage{algorithm}
\usepackage{array}
\usepackage[caption=false,font=footnotesize,labelfont=rm,textfont=rm]{subfig}
\usepackage{textcomp}
\usepackage{stfloats}
\usepackage{url}
\usepackage{verbatim}
\usepackage{graphicx}
\usepackage{subfloat}
\usepackage{cite}
\usepackage{booktabs}
\usepackage{multirow}
\usepackage{color}
\usepackage[colorlinks,bookmarksopen,bookmarksnumbered,citecolor=blue, linkcolor=blue, urlcolor=blue]{hyperref}
\usepackage{cleveref}
\begin{document}

\title{Point Cloud Classification Using Content-based Transformer via Clustering in Feature Space}

\author{
Yahui Liu, Bin Tian, Yisheng Lv, Lingxi Li, and Fei-Yue Wang

\thanks{Y. H. Liu, B. Tian, Y. S. Lv and F.-Y. Wang are with the Institute of Automation,
  Chinese Academy of Sciences, Beijing, China. (e-mail: liuyahui2021@ia.ac.cn; bin.tian@ia.ac.cn;
  yisheng.lv@ia.ac.cn; feiyue@ieee.org).
  
  L. X. Li is with the Purdue School of Engineering and Technology, Indiana University-Purdue University Indianapolis (IUPUI), Indianapolis, USA.
  (e-mail: LL7@iupui.edu).}
}

\markboth{IEEE/CAA JOURNAL OF AUTOMATICA SINICA}
{LIU \MakeLowercase{\textit{et al.}}: Point Cloud Classification Using Content-based Transformer via Clustering in Feature Space}

\maketitle

\begin{abstract}
  Recently, there have been some attempts of Transformer in 3D point cloud classification.  
  In order to reduce computations, most existing methods focus on local spatial attention, but ignore their content and fail to establish relationships between distant but relevant points. 
  To overcome the limitation of local spatial attention, we propose a point content-based Transformer architecture, called PointConT for short.
  It exploits the locality of points in the feature space (content-based), which clusters the sampled points with similar features into the same class and computes
  the self-attention within each class, thus enabling an effective trade-off between capturing long-range dependencies and computational complexity. 
  We further introduce an Inception feature aggregator for point cloud classification, which uses parallel structures to aggregate high-frequency and low-frequency information in each branch separately.
  Extensive experiments show that our PointConT model achieves a remarkable performance on point cloud shape classification.  Especially, our method exhibits
  90.3\% Top-1 accuracy on the hardest setting of ScanObjectNN. 
  Source code of this paper is available at \url{https://github.com/yahuiliu99/PointConT}.
\end{abstract}

\begin{IEEEkeywords}
point cloud classification, local attention, content-based Transformer, feature aggregator, deep learning.
\end{IEEEkeywords}

\section{Introduction}
\IEEEPARstart{3}{D} point cloud analysis has gained tremendous popularity in many fields, including scene understanding\cite{motive,localization,reconstruction},
robotics and self-driving vehicles\cite{lidar,ranseg,track}. 
Compared with 2D images, 3D point clouds can provide sufficient spatial and geometric information, 
but they are not arranged in any particular order. 
Due to its irregular structure, 
the convolutional neural networks cannot be directly applied to point cloud processing, while
Transformer\cite{transformer} architecture is inherently permutation-invariant and natural-suited for point cloud learning.

Recently, some explorations have been made on the Transformer architecture in point cloud analysis \cite{pointtransformer,pct,pointformer,fastpoint,stratified,pvt,unified}.
However, a common downside of these models, the high computational cost, has caught the attention of researchers
and motivated them to consider the trade-off between accuracy and inference speed.
The two main approaches to reduce the computational complexity are downsampling points and local self-attention\cite{pointtransformer,pvt,fastpoint,stratified}.
Points downsampling algorithms, such as farthest point sampling (FPS)\cite{pointnet2}, provide uniform coverage of the entire point cloud.
Local self-attention computes the relationship within a subset of points (patch or cubic window) that is partitioned in 3D space. 
Although local spatial attention significantly improves efficiency, 
it still faces difficulty in capturing interactions among distant but similar points.

\begin{figure}[t]
  \centering
   
   \subfloat[KNN in 3D space]{
    \begin{minipage}[b]{0.225\textwidth}
    \includegraphics[width=\textwidth]{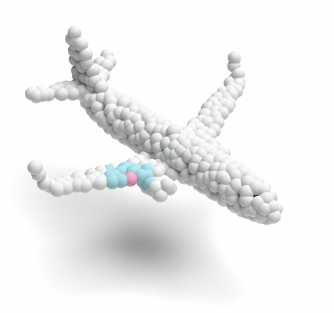}
    \\
    \includegraphics[width=1.02\textwidth]{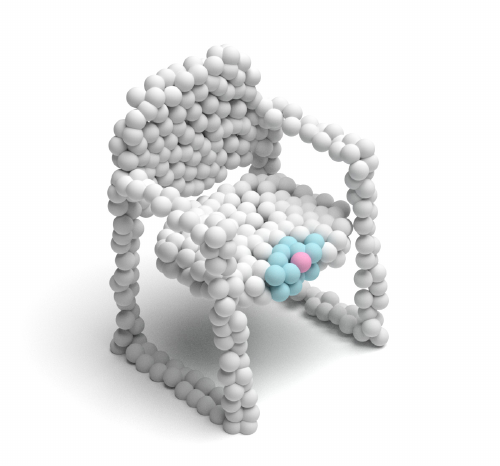}
    \end{minipage}
    } 
    \subfloat[Cluster in feature space (visualize in 3D space)]{
    \begin{minipage}[b]{0.225\textwidth}
    \includegraphics[width=\textwidth]{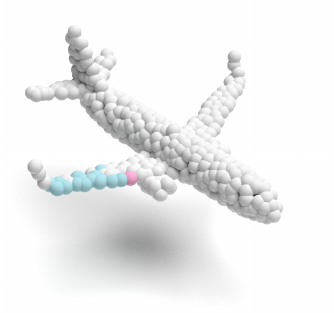}
    \\
    \includegraphics[width=1.02\textwidth]{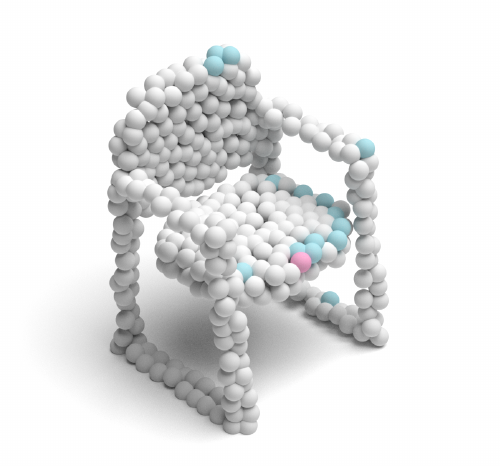}
    \end{minipage}
    }
   \caption{Comparison between 3D space locality and  content-based locality.  The red point denotes the sampled center point, and the blue points denote neighborhood or cluster. In content-based attention, points will be clustered into multiple clusters based on their feature similarity.}
  \label{fig:local}
  \end{figure}

Therefore, we propose a simple yet powerful architecture for point cloud classification, named point content-based Transformer (PointConT), 
\textbf{which exploits local self-attention in the feature space (content-based) instead of the 3D space}, as visualized in \hyperref[fig:local]{Fig.~\ref*{fig:local}}. 
Starting from the content of the point cloud, we cluster the sampled points into classes based on their similarity, 
and compute the self-attention within each class, which preserves the ability of the 
global self-attention mechanism to capture long-range feature dependencies, while significantly reducing computational complexity.
Specifically, it dynamically divides all queries into multiple clusters according to their contents
(\textit{i.e.,} features) in each block, and selects the corresponding keys and values to compute local self-attention. 
The clustering varies accordingly at each stage and each head in the Transformer, adequately reflecting the content dynamics. 
Note that unlike the k-nearest neighborhood (kNN), the clusters are non-overlapping, which further reduces the computational complexity. 

Moreover, we complement the point cloud feature aggregation from a frequency standpoint. 
Recent studies\cite{visvit,iformer} found that max-pooling amplifies high-frequency features while average pooling and Transformer reduce high-frequency components, 
which also accords with the observations in our ablation experiments.
\add{In order to aggregate high-frequency and low-frequency features}, we design an Inception feature aggregator composed of two branches, where the name of ``Inception" is derived from the Inception module\cite{inception,inceptionv4}. 
The high-frequency aggregation branch consists of a max-pooling operation and a residual MLP module, 
while the low-frequency aggregation branch is implemented by an average pooling operation and the content-based Transformer block.

The main contributions of this paper are summarized as follows.

\begin{itemize}
  \item We propose the point content-based Transformer (PointConT) to cluster points according to their content and compute self-attention within each cluster, establishing long-range feature dependencies while significantly reducing computations.
  
  \item We design an Inception feature aggregator for point cloud classification, using parallel structures to aggregate high-frequency and low-frequency information in each branch separately.
  
  \item Experiments show the competitiveness of our model on ModelNet40\cite{ModelNet40} and ScanObjectNN\cite{scanobjectnn} datasets. Extensive ablation studies verify benefits of each component in the PointConT design.
\end{itemize}

\section{Related Work}

\subsection{Point Cloud Processing}
There are mainly two branches of methods for processing the point clouds. One is to convert 
point clouds into a regular grid structure that can be
directly consumed by convolutional neural networks, such as volumetric 
representation\cite{voxelnet,voxnet,octnet} (through voxelization) or images\cite{mvcnn,simpleview} 
(through projection or rendering). 
The other is point-based modeling, 
where the raw point clouds are directly fed into
deep networks without any conversion. This paper focuses on point-based methods.

PointNet is a pioneering work that successfully applies deep architecture on raw point sets \cite{pointnet}. 
It is constructed as a symmetric function using shared \add{multi-layer perceptrons (}MLP\add{)} and max-pooling, 
which guarantees its permutation-invariance. However, PointNet only learns either 
single-point or global features, and thus is limited in capturing interactions among points.
PointNet++ is built on top of the PointNet, which learns hierarchical point cloud features 
and is able to aggregate features in local geometric neighbors using set abstraction \cite{pointnet2} . 

Following them, some works have extended the point-based methods to various local 
aggregation operators. The explorations of local aggregation operators can 
be categorized into three groups: convolution-based\cite{pointcnn,pointconv,pointweb,rscnn,pospool,spidercnn,kpconv,paconv}, graph-based\cite{sonet,dgcnn,specgcn,curvenet}, and attention-based\cite{gdanet,pct,pointtransformer,rpnet,scfnet} methods. 

\subsubsection*{\bf Convolution-based methods}
\cite{rscnn} and \cite{pospool} learn the kernel within a local region through predefined geometric priors. 
Another type of point convolutions, KPConv\cite{kpconv}, relates the weight matrices with predefined kernel points in 3D space.
However, the fixed kernel points may not be optimal for modeling the complicated 3D position variations. 
PAConv constructs a position adaptive convolution operator with a dynamic kernel \cite{paconv}, which assembles basic weight matrices in Weight Bank. 
The assembling coefficients are learned from relative point positions by MLPs.

\subsubsection*{\bf Graph-based methods}
The rise of the graph-based methods began with DGCNN\cite{dgcnn}, which learns on graphs 
dynamically updated at each layer. It proposes a local feature aggregation operator, named EdgeConv, which generates 
edge features that describe the semantic relationships between key points and their neighbors in the feature space. 
Besides, CurveNet explores geometric information by taking guided walks to group contiguous segments of points as curves \cite{curvenet}.

\subsubsection*{\bf Attention-based methods}
Point Cloud Transformer designs offset attention for extracting global features and uses a neighbor embedding strategy to augment local feature representation\cite{pct}. 
Point Transformer proposes a modified Transformer architecture that aggregates local features with vector attention and relative position encoding \cite{pointtransformer}.
Stratified Transformer\cite{stratified},  inspired by Swin Transformer\cite{swin}, partitions the 3D space into non-overlapping cubic windows,
and proposes a stratified strategy for sampling keys.

In addition, PointASNL\cite{pointasnl} leverages non-local network\cite{nonlocal} and adaptive sample module to enhance the long-dependency correlation learning.
Recently, PointNeXt\cite{pointnext} explores more advanced training and data augmentation strategies with the PointNet++ backbone 
to further improve the accuracy and efficiency.

\subsection{Vision Transformer}
In recent years, compared to familiar convolutional networks, 
Transformer architectures have shown great success in 2D images understanding.
Vision Transformer(ViT)\cite{vit} is the first paper that successfully applies a Transformer encoder on images.
It divides an image into non-overlapping patches (tokens), which are then linearly embedded. 
Further, Pyramid ViT (PVT)\cite{pvtv1,pvtv2} proposes a hierarchical structure into Transformer framework.
Transformer in Transformer (TNT)\cite{tnt} extends the ViT baseline with sub-patch-wise attention within patches. 
More recently, Methods of \cite{swin,cswin,neighborhood,shuffle} compute attentions within local windows.
Swin\cite{swin} is a representative approach, which employs two key concepts to improve the original ViT — hierarchical feature maps and shifted window attention.
Beyond image-space hand-crafted window, DGT\cite{dgt} and BOAT\cite{boat} exploit feature-space locality.
DGT\cite{dgt} dynamically divides queries into multiple groups and 
selects the most relevant keys/values for each group to compute the attention. 
BOAT\cite{boat} supplements the existing window-based local attention with the feature space local attention module,
which enables the modeling ability for long-range feature dependencies to be significantly improved.

Although Transformer is highly capable of establishing long-range dependencies, 
recent studies present intuitive visual explanations that Transformer lacks the ability to capture high frequencies that predominantly convey local information \cite{visvit,iformer}.
In other words, Transformer is a low-pass filter. To address this issue, 
Inception Transformer (iFormer)\cite{iformer} designs \add{a channel splitting mechanism 
to adopt parallel convolution path and self-attention path as high-frequency and low-frequency mixers.} 
  
Inspired by the concepts of feature space local attention and features in different frequencies, 
we adopt content-based Transformer and Inception feature aggregator for 3D point cloud classification. 

\begin{figure*}
  \centering
  \includegraphics[width=0.98\textwidth]{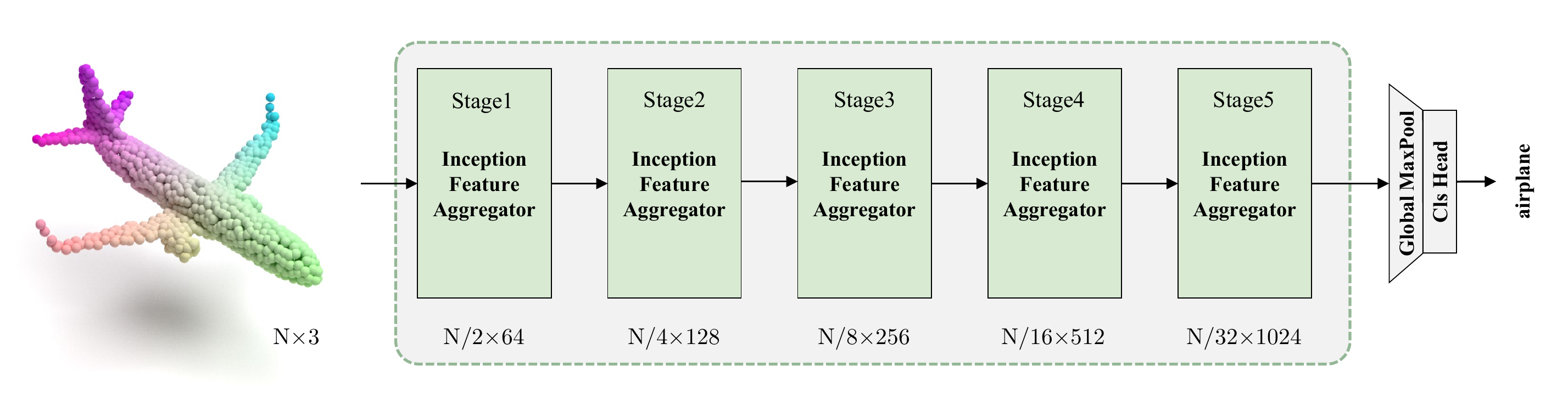}
  \caption{Overall Architecture of Point Content-based Transformer (PointConT). The network is composed of a stack of Inception Feature Aggregator blocks.}
  \label{fig:overall}
\end{figure*}


\section{Methodology}

\subsection{Overall Architecture}

An overview of the proposed PointConT architecture is shown
in \hyperref[fig:overall]{Fig.~\ref*{fig:overall}}.
The backbone structure consists of five hierarchical stages of Inception feature aggregator blocks. 

Given an input point cloud $p\in \mathbb{R}^{N\times 3}$, containing $N$ points in 3-dimensional space.
\add{The “Stage 1” Inception feature aggregator block partitions the point cloud into overlapping patches 
and then embeds the input coordinates into a new feature space (dimension denoted as $C$).}
It halves the number of points and doubles the number of feature dimensions stage by stage. 
Consequently, the output contains $\frac{N}{2^{m}}$ points and $2^{m-1}C$ feature dimensions at the $m$-th stage.
For classification, the final classifier head is a global max-pooling followed by two linear layers.

\begin{figure}
  \centering
  \includegraphics[width=0.45\textwidth]{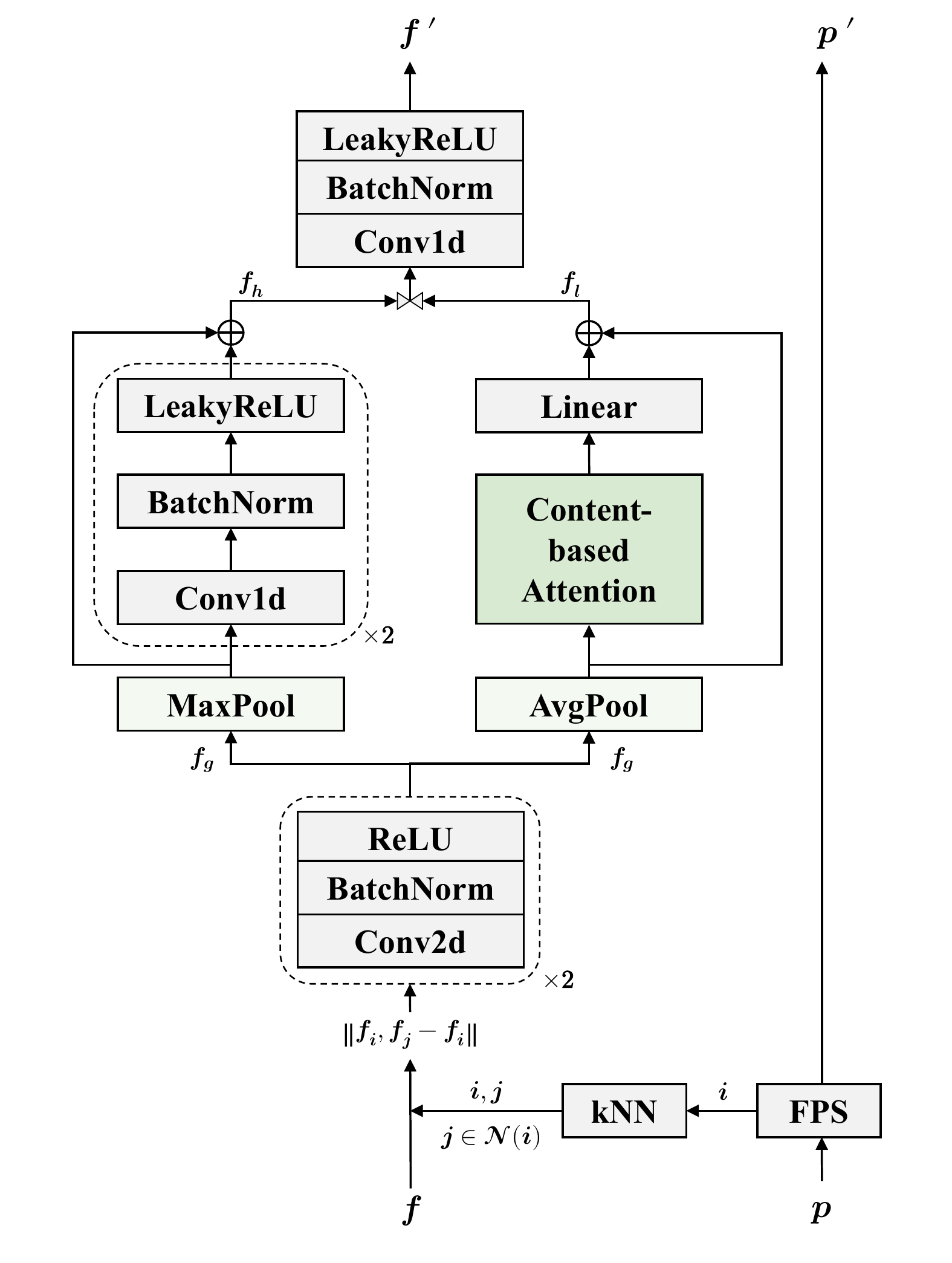}
  \caption{The details of the Inception feature aggregator.}
  \label{fig:inception}
\end{figure}

\subsection{Inception Feature Aggregator}
As shown in \hyperref[fig:inception]{Fig.~\ref*{fig:inception}}, take the $m$-th $(m>1)$ stage as an example. Given the point coordinates $p\in \mathbb{R}^{\frac{N}{2^{m-1}}\times 3}$ and point features $f\in \mathbb{R}^{\frac{N}{2^{m-1}}\times 2^{m-2}C}$ from the last stage, 
the Inception feature aggregator block first downsamples center points at 2 rates from $p$ via FPS, then the kNN algorithm is performed 
to group local patches. \add{We regard $i$ as the center point and $\{j:(i,j)\in \mathcal{N}\}$ as a patch surrounding it}. We further use an EdgeConv designed in DGCNN\cite{dgcnn} shown as Eq. \eqref{eq:edgeconv}, which extracts the relationship between center point feature $f_i\in \mathbb{R}^{\frac{N}{2^{m}}\times 2^{m-2}C}$ and its neighbors $f_j\in \mathbb{R}^{\frac{N}{2^{m}}\times k \times 2^{m-2}C}$ 
($k$ denotes the number of neighbors) within each patch. 
\begin{equation}
    f_g  = \text{MLP}(\|f_i, f_j-f_i\|),\quad     \add{f_g \in \mathbb{R}^{\frac{N}{2^m}\times k\times 2^{m-1}C}}
  \label{eq:edgeconv}
\end{equation}
\add{Where $f_j-f_i$ denotes that $f_j$ minus $f_i$ to obtain the neighboring features relative to the centroid $i$, $\|\cdot\|$ is the concatenation operation 
and MLP is a simple network that includes a point-wise convolutional layer, a batch normalization layer, and an activation function.
}
Note that unlike DGCNN, which defines its kNN in the feature space, we adopt neighbor search in the 3D space.

Next, we propose a mix pooling strategy to aggregate the features of local patches.
In most previous works, max-pooling has been verified as effective in aggregating the local feature\add{s}, 
for the reason that it can capture high frequencies that predominantly convey local information. 
Instead of directly combining max-pooling and Transformer block in a serial manner, in our PointConT, 
we use a parallel structure composed of a high-frequency aggregation branch and a low-frequency aggregation branch.
The max-pooling operation aggregates high-frequency signals, while the average pooling 
operation filters low-frequency representations.

\subsubsection*{\bf High-frequency aggregation branch} 
This branch can be defined as
\begin{equation}
  f_h = \text{ResMLP}(\text{MaxPool}(f_g)),\quad     \add{f_h \in \mathbb{R}^{\frac{N}{2^m}\times 2^{m-1}C}}
\end{equation}
where $\text{MaxPool}$ and $\text{ResMLP}$ denote max-pooling operation and residual MLP block, respectively.

\subsubsection*{\bf Low-frequency aggregation branch}
We  simply utilize an average pooling layer (AvgPool) before the content-based Transformer (ConT),
and this design allows the content-based Transformer to focus on embedding low-frequency information. 
This branch can be defined as
\begin{equation}
  f_l = \text{ConT}(\text{AvgPool}(f_g)),\quad     \add{f_l \in \mathbb{R}^{\frac{N}{2^m}\times 2^{m-1}C}}
\end{equation}

In the end, we concatenate the features from the high-frequency aggregation branch and the low-frequency aggregation branch, 
and then feed them to an MLP block as the Inception aggregator output \add{features} $f'$.
\begin{equation}
  f' = \text{MLP}(\|f_h,f_l\|), \quad     f' \in \mathbb{R}^{\frac{N}{2^m}\times 2^{m-1}C}
\end{equation}

\begin{figure}
  \centering
  \includegraphics[width=0.5\textwidth]{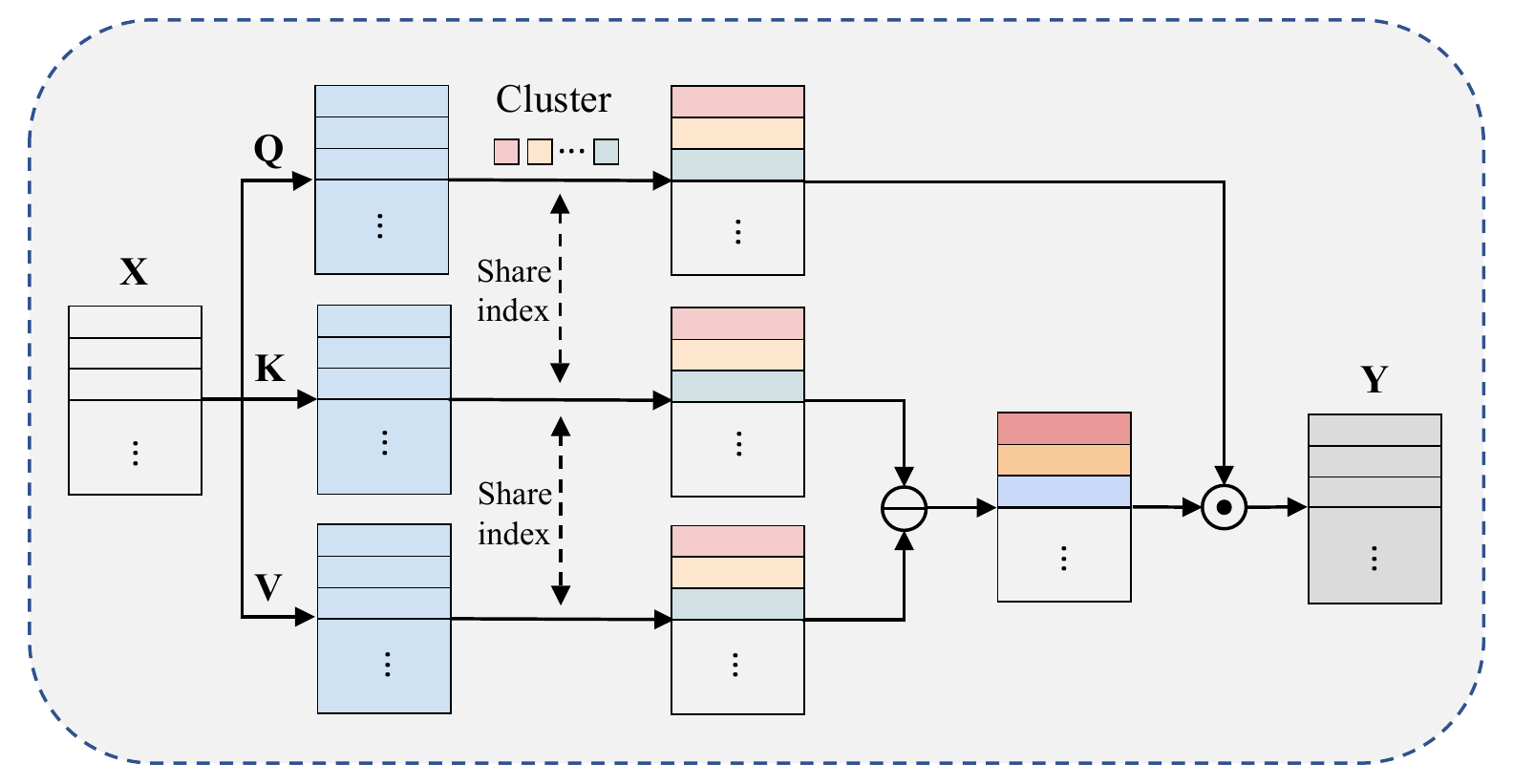}
  \caption{Illustration of content-based attention. It can dynamically cluster all queries into multiple groups and \add{compute} the self-attention within each group.}
  \label{fig:attention}
\end{figure}

\subsection{Content-based Transformer}
Differently from Point Transformer\cite{pointtransformer}, which computes self-attention among 
local spatial neighbors, we propose a content-based attention, as visualized in \hyperref[fig:attention]{Fig.~\ref*{fig:attention}}. 
It dynamically divides all queries into multiple clusters according to their content
(\textit{i.e.,} features) at each block, and selects the corresponding keys and values to compute local self-attention. 

Let $X\in \mathbb{R}^{S \times d}$ ($d$ is the feature space dimension, $S$ denotes the length of features) be a set of feature vectors. 
Furthermore, we get embeddings $Q=XW_Q$, $K=XW_K$ and $V=XW_V$ to represent the queries, 
keys and values, respectively. 

Then we use the clustering algorithm so that the queries are scattered in different clusters.
K-means clustering algorithm is a classic method for clustering problems. 
However, K-means clustering generally enables each cluster to contain varying numbers of queries, 
and therefore this algorithm cannot be implemented in a parallel way by using GPUs. 
To address this issue, we refer to the balanced binary clustering algorithm proposed in BOAT\cite{boat}, 
which equally divides a set of queries into two clusters hierarchically. 

Through clustering, the queries $Q$ are grouped into $L$ subsets $\{Q_i\}_{i=1}^{L}$, where each subset has equal size $|Q_i|=\frac{S}{L}$.
Subsequently, keys $K$ and values $V$ are separated into $\{K_i\}_{i=1}^{L}$ and $\{V_i\}_{i=1}^{L}$ by the same index. 
The self-attention (SA) in each subset is formulated as
\begin{equation}
  Y_i = \text{SA}(Q_i, K_i, V_i)
\end{equation}
where $Y_i$ is the output of each subset. 
Lastly, all subsets $\{Y_i\}_{i=1}^L$ are merged into the output $Y\in \mathbb{R}^{S \times d}$ in keeping with their original order.

\begin{figure}[t]
  \centering
   \subfloat[Head 1]{
    \includegraphics[width=0.2\textwidth]{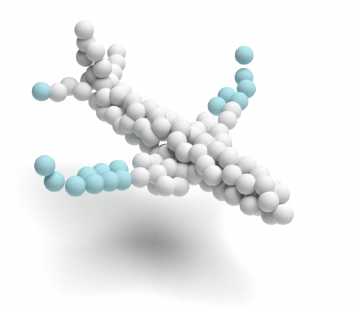}
    }
  \subfloat[Head 2]{
    \includegraphics[width=0.2\textwidth]{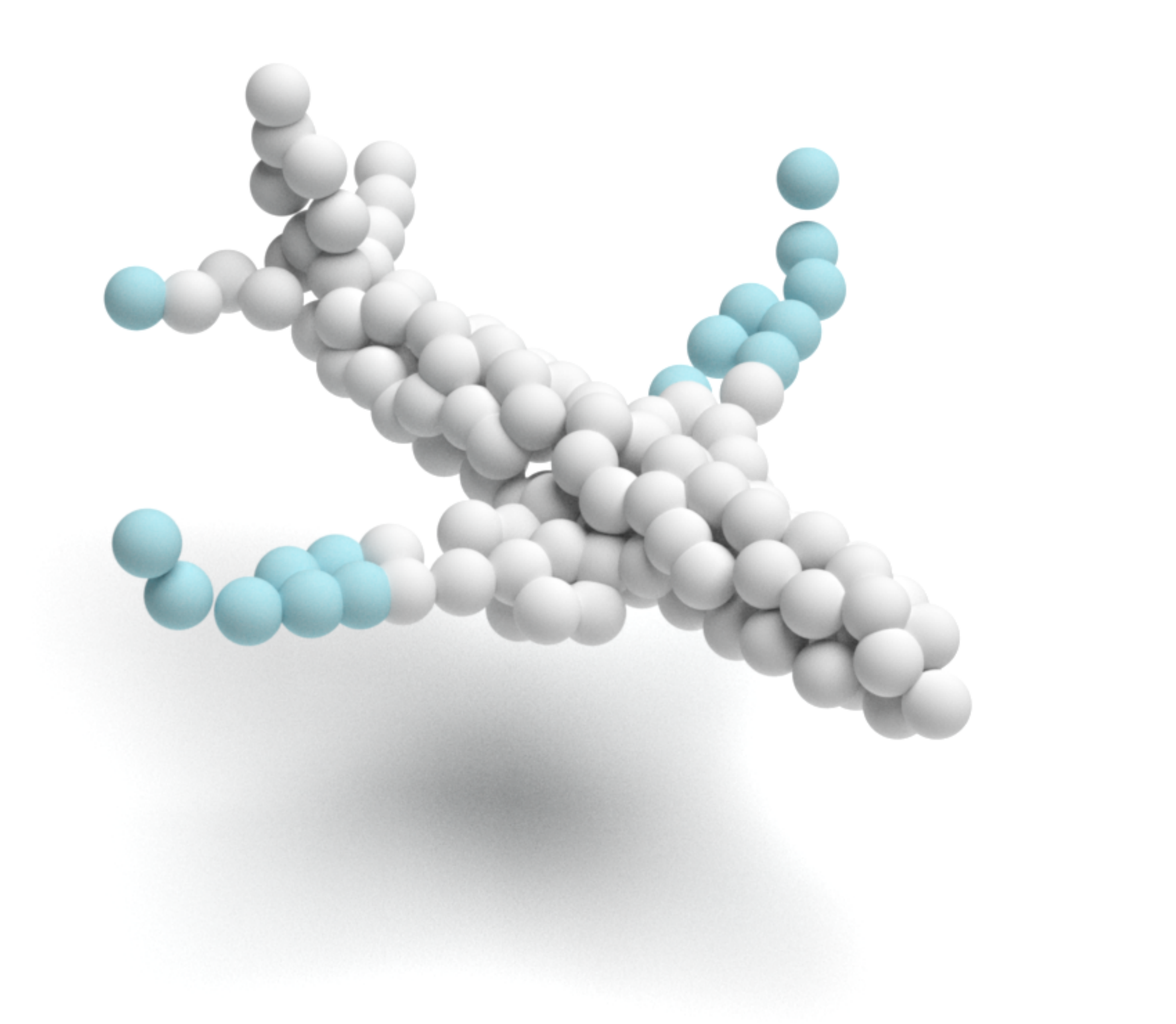}
    }

    \subfloat[Head 3]{
    \includegraphics[width=0.2\textwidth]{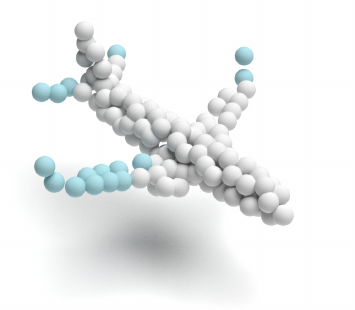}
    }
  \subfloat[Head 4]{
    \includegraphics[width=0.2\textwidth]{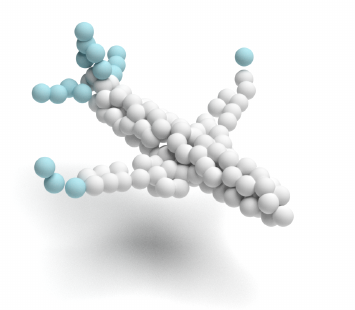}
    }  
   \caption{Visualization of points clustering in different head\add{s}.}
   \label{fig:multi-head}
\end{figure}

Multi-head configuration is a standard practice in Transformer, we expand multiple heads and each head performs query/key/value embeddings and clustering independently.
This setting brings clustering diversity to a great extent, as visualized in \hyperref[fig:multi-head]{Fig.~\ref*{fig:multi-head}}.

\subsubsection*{\bf Hierarchical binary clustering} 
Similarly to K-means clustering that the cluster assignment relies on the distance between all cluster centroids and each sample, 
our binary clustering starts with a random division of queries $Q=\{q_i\}_{i=1}^{S}$  into two clusters and then calculates the two cluster centroids, 
denoted as $c_1$ and $c_2$, respectively. After that, we compute the distance ratio to perform the hard assignment. Above operations can be summarized as
\begin{equation}
  \left\{\,
  \begin{aligned}
    & c_1 = \frac{\sum \{q_i\}_{i=1}^{\frac{S}{2}}}{\frac{S}{2}}, \quad c_2 = \frac{\sum \{q_i\}_{i=\frac{S}{2}+1}^{S}}{\frac{S}{2}} \\
    & r_{i} =\frac{dist\left(q_{i}, c_{1}\right)}{dist\left(q_{i}, c_{2}\right)}, \forall i \in[1,S]\\
    & [i_1, \cdots, i_S] = \text{argsort}(\{r_i\}_{i=1}^{S}) \\
    & \mathcal{C}_1=\{q_{i_j}\}_{j = 1}^{\frac{S}{2}}, \quad \mathcal{C}_2=\{q_{i_j}\}_{j = \frac{S}{2}+1}^{S}
  \end{aligned}
  \right.
\end{equation}
where $dist$ means the Euclidean distance in feature space, $\mathcal{C}_1$ and $\mathcal{C}_2$ represent the two equal size clusters through balanced binary clustering. 
After performing $n$ iterations ($n=\log_2L$) of binary clustering, we obtain $L$ subsets with the same size.

\subsubsection*{\bf Choice of SA\label{sec:SA}} 
In Point Transformer, the choice of the self-attention has a crucial influence on the properties of Transformer block. 
One choice of SA is the standard scalar attention as 
\begin{equation}
  \text{SA} = \text{Softmax}(\frac{QK^T}{\sqrt{d}})V
\end{equation}

Another choice of SA is the vector attention\cite{pointtransformer,vectorattention} that we adopt in this paper
\begin{equation}
  \text{SA} = \text{Softmax}(\frac{Q-K}{\sqrt{d}})\odot V
\end{equation}

\subsubsection*{\bf Complexity Analysis} Given a set of feature vectors with a size $S\times d$, for standard multi-head self-attention (MSA), 
the computational complexity of a local MSA module is $\Omega(S\times(4kd^{2}+2k^{2}d))=4 Skd^{2}+2Sk^{2} d$, where $k$ is the number of points in a local neighborhood.
Unlike scalar attention, vector attention further reduces the complexity to $\Omega(4 Skd^{2}+2Skd)$. 
In our PointConT, hierarchical binary clustering algorithm divides the feature vectors in a non-overlapping manner, dramatically reducing the complexity to $\Omega(4 Sd^{2}+2Sd)$.

\begin{equation}
  \left\{\,
  \begin{aligned}
    &\Omega(\mathrm{MSA_{Local}})=4 Skd^{2}+2Sk^{2} d \\
    &\Omega(\mathrm{MSA_{PointTrans.}})=4 Skd^{2}+2Skd  \\
    &\Omega(\mathrm{MSA_{PointConT}})=4 Sd^{2}+2Sd \\
    \end{aligned}
  \right.
\end{equation}


\section{Experiments}
In this section, we show experimental results of the proposed model on the shape classification task. All the experiments 
are performed on one Tesla V100 GPU.

\noindent\textbf{Implementation details.}
We \add{implement} the PointConT in PyTorch framework and \add{train} the network using the SGD optimizer (momentum and weight decay set to 0.9 and 0.0001, respectively),
cosine learning rate schedule starting at 0.001 (warm up steps set to 10), 
and cross-entropy with label smoothing. 
We \add{fix} the random seed in all experiments to eliminate the influence of randomness.

For shape classification training, we only \add{use} 1024 uniformly sampled points as network inputs. Moreover, we \add{use} RSMix\cite{rsmix} in addition to random scaling and translation as data augmentation. 
We \add{train} PointConT on ModelNet40 and ScanObjectNN with a batch size of 32 and 64 for 300 and 400 epochs, respectively.
For testing, batch size \add{is} set to 16 and 32 on ModelNet40 and ScanObjectNN, respectively. 

\subsection{Classification on ModelNet40}

We evaluate the model on the ModelNet40\cite{ModelNet40} shape classification benchmark. There are 
12,308 computer-aided design (CAD) models of point clouds from 40 common categories. 
The dataset is divided as 9840 models for training and 2468 models for testing.

\begin{figure*}
  \centering
  \includegraphics[width=0.98\textwidth]{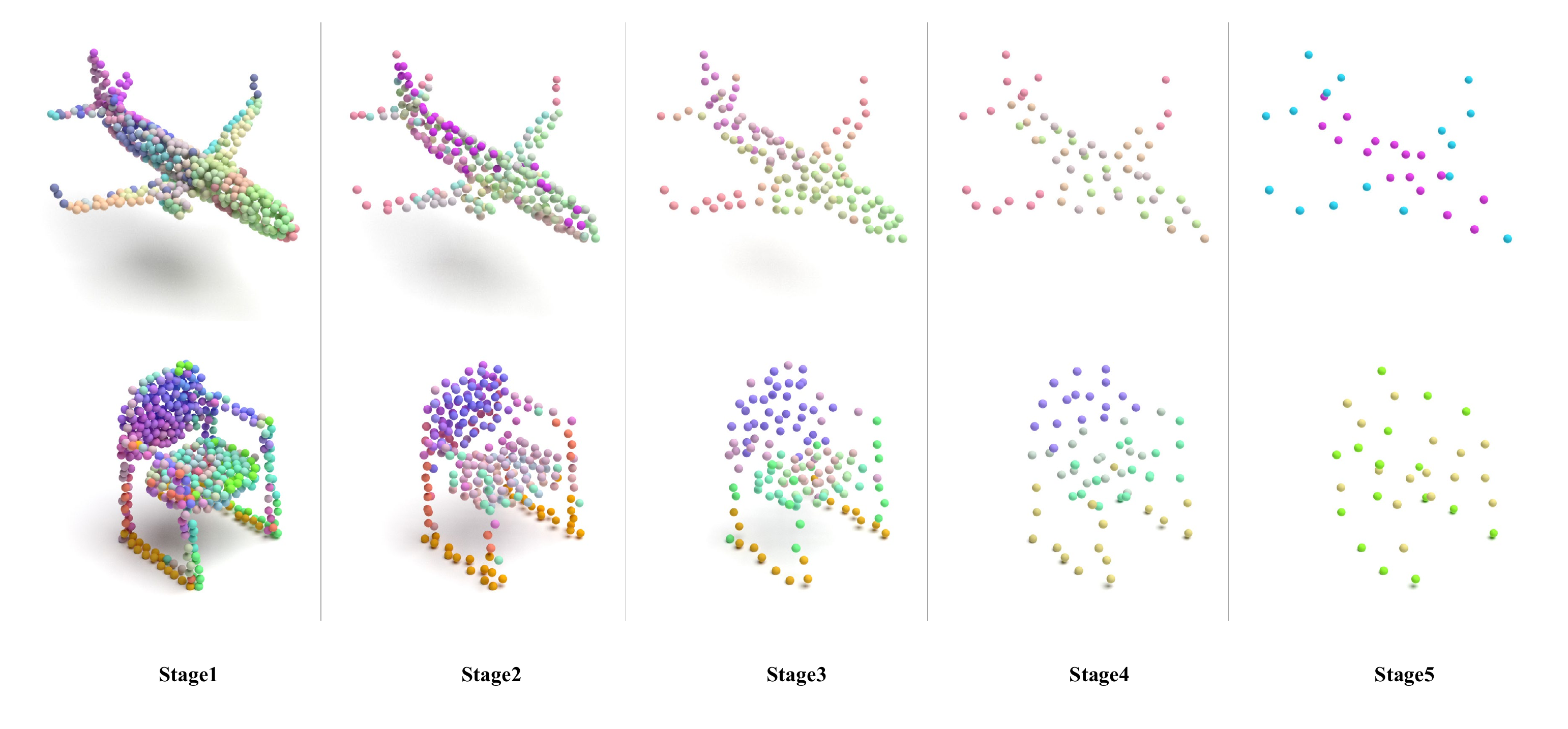}
  \caption{Visualizations of clustering results at each stage. The points of the same cluster are plotted with the same color. Different clusters are distinguished by random colors.}
  \label{fig:cluster}
\end{figure*}

The results are presented in \hyperref[Table:cls]{Table~\ref*{Table:cls}}.  
The overall accuracy of PointConT on ModelNet40 is 93.5\%, which is a competitive result in attention-based models.
Besides, our PointConT presents a high inference speed (166 samples/second in training and 279 samples/second in testing),
which is 3.5$\times$ faster than the original PointMLP\cite{pointmlp} paper and 1.4$\times$ faster than the lightweight version PointMLP-elite.
We visualize the clustering results at each stage in \hyperref[fig:cluster]{Fig.~\ref*{fig:cluster}}. The clusters are able to cover long-range dependencies.

\begin{table}[h]
  \caption{Shape Classification Results on the ModelNet40 Dataset.\protect\\
  \add{OA: Overall accuracy.}}
    \label{Table:cls}
    \begin{center}
    \begin{tabular}{c|l|c|c}
    \toprule
    Method & Model & Input & OA(\%) \\
    \midrule
    \multirow{4}{*}{MLP} & PointNet\cite{pointnet} & 1K points+normal & 89.2\\
    ~ & PointNet++\cite{pointnet2} & 1K points & 90.7\\
    ~ & PointNet++\cite{pointnet2} & 5K points+normal & 91.9\\
    ~ & PointMLP\cite{pointmlp} & 1K points & \bfseries{94.1}\\ 
    \midrule
    \multirow{7}{*}{Conv} & PointCNN\cite{pointcnn} & 1K points & 92.5\\
    ~ & PointWeb\cite{pointweb} & 1K points+normal & 92.3\\
    ~ & PointConv\cite{pointconv} & 1K points+normal & 92.5\\
    ~ & RS-CNN\cite{rscnn}  & 1K points & 92.9\\
    ~ & KPConv\cite{kpconv} & 1K points & 92.9\\
    ~ & PosPool\cite{pospool} & 5K points & 93.2\\
    ~ & PAConv\cite{paconv}  & 1K points & 93.6\\
    \midrule
    \multirow{2}{*}{Graph} & DGCNN\cite{dgcnn} & 1K points & 92.9\\
    ~ & CurveNet\cite{curvenet} & 1K points & 93.8\\
    \midrule
    Non-Local & PointASNL\cite{pointasnl} & 1K points & 92.9\\
    \midrule
    \multirow{5}{*}{Attention} & GDANet\cite{gdanet}  & 1K points & 93.4\\
    ~ & PCT\cite{pct} & 1K points & 93.2\\
    ~ & Point Trans.\cite{pointtransformer} & 1K points & 93.7\\ 
    ~ & PointConT(ours) & 1K points & 93.5\\
    \bottomrule
    \end{tabular}
    \end{center}
\end{table}

\subsection{Classification on ScanObjectNN}
We furthermore perform experiments on a recent real-world point cloud classification dataset — ScanObjectNN\cite{scanobjectnn}, 
which consists of 15k objects from 15 categories. We only report the heavy permutations from rigid
transformations PB\_T50\_RS dataset.
Unlike sampled virtual point clouds in ModelNet40, objects in ScanObjectNN are obtained from real-world scans. 
Therefore, the point clouds in ScanObjectNN are noisy (background, occlusions) and not axis-aligned, 
which brings a significant challenge to existing point cloud analysis methods.

\hyperref[Table:cls2]{Table~\ref*{Table:cls2}} shows the classification results on ScanObjectNN. PointConT outperforms prior models with 88.0\% overall accuracy 
without voting\cite{rscnn} and reaches Top-1 90.3\%  when averages 10 prediction votes.  
This suggests that the PointConT is effective in real world point clouds. 

\begin{table}[h]
  \caption{Shape Classification Results on the ScanObjectNN Dataset. \protect\\
  $\quad *$ denotes method evaluated with voting strategy\cite{rscnn}. \protect\\
  \add{mAcc: mean class accuracy; OA: overall accuracy.}}
    \label{Table:cls2}
  \begin{center}
    \begin{tabular}{l|c|c}
    \toprule
    Model & mAcc(\%) & OA(\%) \\
    \midrule
    PointNet\cite{pointnet}  & 63.4 & 68.2\\
    SpiderCNN\cite{spidercnn} & 69.8 & 73.7\\
    PointNet++\cite{pointnet2} & 75.4 & 77.9\\
    PointCNN\cite{pointcnn} & 75.1 & 78.5\\
    DGCNN\cite{dgcnn} & 73.6 & 78.1\\
    GBNet\cite{GBNet}  & 77.8 & 80.5\\
    SimpleView\cite{simpleview}  & - & 80.5\\
    PRANet\cite{pranet} & 79.1 & 82.1\\
    Point-BERT\cite{pointbert}  & - & 83.1\\
    Point-MAE\cite{pointmae} & - & 85.2\\
    PointMLP\cite{pointmlp} & 84.4 & 85.7\\
    PointNeXt\cite{pointnext} & 86.8 & 88.2\\
    \midrule
    PointConT(ours) & 86.0 & 88.0\\
    PointConT(ours)* & \bfseries{88.5} & \bfseries{90.3}\\
    \bottomrule
    \end{tabular}
    \end{center}
\end{table}




\subsection{Ablation Study}
We perform ablation studies for the key designs of our methods on the shape classification task. 
All experiments are conducted under the same training settings.

\subsubsection*{\bf Component ablation study}
\hyperref[Table:Ablation study]{Table~\ref*{Table:Ablation study}} reports the classification results of removing each component in PointConT. 
Comparing Exp.III and Exp.IV, we notice that with the content-based Transformer, the model improves with 0.6\% on ModelNet40 and 1.7\% on ScanObjectNN. 
This demonstrates that the content Transformer can enhance the representation power of point clouds.
Remarkably, the result of Exp.V drops a lot. In the absence of the average pooling, 
Exp.V means that the content-based Transformer follows after max-pooling and residual MLP in a serial manner, 
which indicates that the mix pooling strategy plays an important role in PointConT.
Observably, by combining all these components, we obtain the best results on ModelNet40 and ScanObjectNN, 
which implies the effectiveness of content-based Transformer and Inception feature aggregator in point cloud classification.

\begin{table}[h]
  \caption{Ablation study. \textbf{MP}: Max-pooling; \textbf{Res}: Residual MLP; \textbf{AP}: Average pooling; \textbf{ConT}: Content-based Transformer. 
  \protect\\ Metric: OA(\%).}
  \label{Table:Ablation study}
  \centering
  \begin{tabular}{c|cccc|cc}
    \toprule
  ID & MP     & Res      & AP      & ConT         & ModelNet40    & ScanObjectNN  \\ \midrule
  I & $\checkmark$ &              &              &              & 93.2          & 86.5          \\
  II & $\checkmark$ & $\checkmark$ &              &              & 93.1          & 87.3          \\
  III & $\checkmark$ & $\checkmark$ & $\checkmark$ &              & 92.9          & 86.3          \\ 
  IV & $\checkmark$ & $\checkmark$ & $\checkmark$ & $\checkmark$ & \textbf{93.5} & \textbf{88.0} \\ \midrule
  V & $\checkmark$ & $\checkmark$ &              & $\checkmark$ & 92.8          & 87.2          \\
  VI & $\checkmark$ &              & $\checkmark$ & $\checkmark$ & 92.8          & 87.2          \\
  VII &             &              & $\checkmark$ & $\checkmark$ & 93.0          & 81.0          \\
  \bottomrule
\end{tabular}
\end{table}

\add{\subsubsection*{\bf The number of stages}
In \hyperref[Table:The number of stages]{Table~\ref*{Table:The number of stages}}, we ablate different number of stages in PointConT. 
We gradually increase the depth of PointConT on ModelNet40 and ScanObjectNN datasets to test the effectiveness of greater depth.
We find that the stage number of 5 is sufficient for full exploitation.
A deeper model will bring redundant information and performance decline.

\begin{table}[h]
  \caption{\add{Ablation study: The number of stages.}}
  \label{Table:The number of stages}
  \centering
  \begin{tabular}{c|cc}
    \toprule
    \add{The number of stages}   & \add{ModelNet40}    & \add{ScanObjectNN}  \\ \midrule
  \add{3}  & \add{91.9}    & \add{84.9}          \\
  \add{4}  & \add{92.7}    & \add{86.8}          \\
  \add{\textbf{5}}   & \add{\textbf{93.5}} & \add{\textbf{88.0}} \\ 
  \add{6}  & \add{92.8}    & \add{86.3}          \\
  \bottomrule
\end{tabular}
\end{table}
}

\subsubsection*{\bf Local cluster size}
We investigate the setting of the local cluster size and show the result in \hyperref[Table:Local cluster size]{Table~\ref*{Table:Local cluster size}}.
The best performance is achieved when the local cluster size is set to 16.  

\begin{table}[h]
  \caption{Ablation study: Local cluster size.}
  \label{Table:Local cluster size}
  \centering
  \begin{tabular}{c|cc}
    \toprule
  Local cluster size   & ModelNet40    & ScanObjectNN  \\ \midrule
  \add{8}  & \add{92.9}    & \add{87.7}          \\
  \textbf{16}   & \textbf{93.5} & \textbf{88.0} \\ 
  32       & 92.8          & 87.7          \\
  \bottomrule
\end{tabular}
\end{table}

\subsubsection*{\bf Similarity metric}
We compare two important measures of similarity for clustering: 
the cosine similarity and the Euclidean distance. 
The cosine similarity is proportional to the dot product of two vectors. Hence, 
vectors with a high cosine similarity lied in the close direction from the origin, 
while the Euclidean distance corresponds to the L2-norm of a difference between vectors. 
Vectors with a small Euclidean distance are located in the close region of a vector space.
The result in \hyperref[Table:Similarity metric]{Table~\ref*{Table:Similarity metric}} shows that clustering according to Euclidean distance
is better than cosine similarity in the classification task.

\begin{table}[h]
  \centering
  \caption{Ablation study: Similarity metric.}
  \label{Table:Similarity metric}
  \begin{tabular}{c|cc}
    \toprule
  Similarity metric   & ModelNet40    & ScanObjectNN  \\ \midrule
  cosine similarity      & 92.9          & 87.8          \\
  \textbf{Euclidean distance}   & \textbf{93.5} & \textbf{88.0} \\ 
  \bottomrule
\end{tabular}
\end{table}

\subsubsection*{\bf Attention type}
Finally,  we compare the scalar attention and the vector attention \add{introduced in \hyperref[sec:SA]{Sec.~\ref*{sec:SA}}}
The results are shown in \hyperref[Table:Attention type]{Table~\ref*{Table:Attention type}}. 
It is obvious that the attention module is more effective than the no-attention, 
and vector attention slightly outperforms scalar attention.  
As described in Point Transformer, vector attention supports adaptive modulation of individual feature channels, 
rather than just the entire feature vector, which can be beneficial in 3D point cloud analysis.

\begin{table}[h]
  \caption{Ablation study: Attention type.}
  \label{Table:Attention type}
  \centering
  \begin{tabular}{c|cc}
    \toprule
  Attention type     & ModelNet40    & ScanObjectNN  \\ \midrule
  no attention       & 92.9          & 86.3          \\
  scalar attention   & 92.9          & 87.9          \\
  \textbf{vector attention}   & \textbf{93.5} & \textbf{88.0} \\ 
  \bottomrule
\end{tabular}
\end{table}

\section{Conclusion}
In this paper, we propose Point Content-based Transformer (PointConT), a simple yet powerful architecture, 
adopting content-based Transformer, which clusters the sampled points with similar features into the same class and computes the self-attention within each class. 
Content-based Transformer can establish long-range feature dependencies compared to local spatial attention.
Moreover, we design an Inception feature aggregator to combine high-frequency and low-frequency information in a parallel manner. 
The max-pooling operation aggregates high-frequency signals, while the average pooling operation and Transformer filter low-frequency representations.
We hope that this study will provide valuable insights into the point cloud Transformer designs.

\add{It is noticed that the balanced clustering algorithm generates clusters with the same size, which limits generality and flexibility of the proposed PointConT.
Advanced clustering approaches and CUDA can be used to implement cluster-wise matrix multiplication in future work.
}



\bibliographystyle{IEEEtran}
\bibliography{IEEEabrv,egbib}

\vfill

\end{document}